\documentclass[10pt,a4paper,twocolumn]{article}

\usepackage[utf8]{inputenc}
\usepackage[T1]{fontenc}
\usepackage{amsmath}
\usepackage{amsfonts}
\usepackage{amssymb}
\usepackage{graphicx, graphics}
\usepackage{bm} % for bold symbols
\usepackage[margin=0.9in]{geometry}
\graphicspath{{./grf/}}

\author{Mohsen Farhadloo, PhD \\John Molson Scool of Business, Concordia University\\ \texttt{mohsen.farhadloo@concordia.ca}}

\title{Twitter Sentiment on Affordable Care Act using Score Embedding}
\begin{document}
\maketitle
%%%%%%%%%%%%%%%%%%%%%%%%%%%%%%%%%%%%%%%%%%%%%%%%%%
\begin{abstract}
	In this paper we introduce \textit{score embedding}, a neural network based model to learn interpretable vector representations for words. Score embedding is a supervised method that takes advantage of the labeled training data and the neural network architecture to learn interpretable representations for words. Health care has been a controversial issue between political parties in the United States. In this paper we use the discussions on Twitter regarding different issues of affordable care act to identify the public opinion about the existing health care plans using the proposed score embedding. Our results indicate our approach effectively incorporates the sentiment information and outperforms or is at least comparable to the state-of-the-art methods and the negative sentiment towards ``TrumpCare'' was consistently greater than neutral and positive sentiment over time.
\end{abstract}

%%%%%%%%%%%%%%%%%%%%%%%%%%%%%%%%%%%%%%%%%%%%%%%%%%
\section{Introduction}
Sentiment analysis as a type of text categorization is the task of identifying the sentiment orientation of documents written in natural language which assigns one of the pre-defined sentiment categories into a whole document or pieces of the document such as phrases or sentences \cite{Pang_08-opinion,Farhadloo_16-fundamentals}. Many studies used binary classification and reported high performance \cite{Maas_11-learning, Tang_14-sswe, Paul_17-compass} and some studies have observed that the performance of the categorization reduces as the number of sentiment categories increases \cite{Bermingham_11-using, Le_14-distributed, BravoMarquez_15-unlabelled, Hamilton_16-inducing}. 

Bag-Of-Words (BOW), a standard approach for text categorization, represents a document by a vector that indicates the words that appear in the document. Although BOW is a widely used approach, it is well-known that it does not preserve the order of the words in the document. The resulting representation suffers from sparsity as not many of the words in the vocabulary occur in a document.

Augmenting the unigrams with n-grams is the traditional approach to preserve the order of the occurrence of words in a documents, but it has not been always effective. Recently, neural network models, such as convolutional neural network \cite{Lecun_98-gradient}, have been employed to exploit the word order and the internal structure of documents in text categorization \cite{Kim14-convolutional,JohnsonZhang14-effective}. 

To address the sparsity issue of BOW representations, word vectors have been introduced to find a real-valued low-dimensional vector representation for words. Different methods have been proposed to calculate the vector representations of single words which have been widely used as features in various text categorization tasks \cite{Mikolov_13-efficient, Weston_14-tagspace}. Many of the popular word vectors are learned in an unsupervised manner over huge corpora. 

In the existing embedding techniques, the dimension which words are embedded is a hyper-parameter of the models and is not known a priori. Additionally, the initialization of the embeddings has an impact on the representations. The quality of word vectors has been studied through similarity tasks, but the meanings and interpretations of the dimensions in the vector space have mainly not been discussed. 

%%%%% Contributions
% Slightly tuned
In this paper we introduce score embedding as a neural network based model for text categorization. Score embedding as a supervised approach takes advantage of the labeled training samples in two ways. First, it leverages the labeled training data to learn pre-trained embeddings in a way that does not require massive amount of unlabeled data. Second, it uses convolutional layers to learn meaningful and interpretable vector representation for words by fine-tuning the representations using the output class labels. In score embedding, the dimension of the embedding is the number of available categories in the particular categorization task and therefore is known a priori. This is an advantage compared to other embedding methods where the embedding dimension is a hyper-parameter of the model. Each dimension in the score embedding space can be interpreted as the association of the given word with the corresponding category. Our approach is appropriate for supervised text categorization tasks and hence takes advantage of the available labeled training data. Score embedding, compared to the existing unsupervised embedding techniques, does not require a massive amount of unlabeled data, but requires labeled training data and is capable of fine tuning the learned scores using the neural network architecture. 

We examined the performance of the proposed score embedding on two tasks. First, we have used our approach to identify the public opinions towards the health care plans in the United States using the on going discussions on Twitter about different issues related to affordable care act. Score embedding was compared with other existing methods including classic supervised classifications and word embedding methods based on neural networks. Second, to be able to compare the effectiveness of score embedding with the existing state-of-the-art methods, we have conducted fine-grained sentiment categorization using the Stanford Sentiment Treebank \cite{Socher_13-recursive} as a benchmark.  

%%%%%%%%%%%%%%%%%%%%%%%%%%%%%%%%%%%%%%%%%%%%%%%%%%
\section{Related work}
%% write about re-embeddings
%Our work is related to existing studies promoting the idea that the representation of words heavily relies on the applications or tasks in which it is used \cite{Labutov_13-reEmbedding}. Particularly for the sentiment analysis task, the existing unsupervised word vector representations are not able to effectively capture the sentiment since they are not incorporating any sentiment information in their learning process and they are designed to capture the syntactic contexts. Our work is related to \cite{Maas_11-learning, Tang_14-sswe} which have tried to address the shortcomings of the existing unsupervised word vectors methods for the task of sentiment analysis by incorporating the sentiment related supervision.

Our work is based on the idea that the representation of words heavily relies on the applications or tasks in which it is used \cite{Labutov_13-reEmbedding}. Particularly for sentiment analysis, existing unsupervised word vector representations are not able to effectively capture the sentiment since they do not incorporate any sentiment information in their learning process and are designed to capture the syntactic contexts. There are studies that have tried to propose models for composition to improve the performance of semantic word vectors for tasks such as sentiment categorization \cite{Yessenalina_11-compositional,Socher_12-semantic,Socher_13-recursive}. These supervised models are designed to capture the meaning of longer phrases other than single words and are based on the parse tree of the input phrase. There are studies that have used supervised techniques to learn vector representations \cite{Weston_14-tagspace,Gao_15-modeling} for words and documents for particular applications such as hashtag prediction and interestingness of web pages. However, our work is related to \cite{Maas_11-learning, Tang_14-sswe} which have tried to address the shortcomings of the existing unsupervised word vectors methods for the task of sentiment analysis by incorporating the sentiment related information in their learning process. Unlike \cite{Maas_11-learning} that follow the probabilistic topic models for documents in a semi-supervised approach, we propose a supervised neural network architecture to incorporate the sentiment information. Unlike \cite{Tang_14-sswe} that learn the sentiment specific embedding with noisy distance supervision, we use the manually labeled data to learn our model.

The proposed method of score embedding is based on the idea of word embedding and score representation. Hence, here we give a brief overview to these techniques.

%There are studies that have used supervised techniques to learn vector representations \cite{Weston_14-tagspace,Gao_15-modeling} for words and documents for particular applications such as hashtage prediction and interestingness of webpages. %\cite{Weston_14-tagspace} argues that the abundance of hashtags in real posts provides a huge labeled dataset and can be used as a stronger semantic guidance than the unsupervised methods and uses a convolutional neural architecture for learning low-dimensional vector representations which are good for hashtag prediction. In \cite{Gao_15-modeling} feature vectors in a latent space are learned for the source-target document pairs in such a way that the distance between source documents and their corresponding interesting targets are minimized in that space. 

%Also similar to our paper, \cite{Wong_15-twitter} has examined whether the Twitter sentiment (the positivity or negativity of tweets) can be used to predict the state-level Affordable Care Act (ACA) marketplace enrollments. In this study only positive and negative sentiments have been considered and lexicon-based approach has been used to identify the sentiments although it is known that the word's sentiment depends on the domain in which it is used \cite{Hamilton_16-inducing}.

%%%%%%%%%%%%%%%%%%%%%%%%%%%
\subsection{Word embedding}
%In computational linguistics word embeddings (distributed representations) are discussed in the distributional semantics research area where the underlying idea is that a word is characterized by the company it keeps \cite{Firth57-synopsis}. The objective in distributional semantics is to quantify the semantic similarities among linguistic items based on their distributional properties in a large sample of language data. In distributional semantics, vector space models have been used to map (embed) each word from a space with one dimension per word to a continuous space (real-valued) with much lower dimensions. 

In the field of distributional semantics, the underlying idea behind word embeddings is that a word is characterized by the company it keeps \cite{Firth57-synopsis}. The objective in distributional semantics is to quantify the semantic similarities among linguistic items based on their distributional properties in a large sample of language data. Vector space models have been used to map (embed) each word from a space with one dimension per word to a continuous space (real-valued) with fewer dimensions. 

Continuous Bag-Of-Words (CBOW), Skip-grams and GloVe are among the most popular word embedding techniques.
%Neural network based language models (NNLM) \cite{Bengio_03-neural,Collobert_11-natural} consist of three main blocks: the \textit{embedding block} which generates vector representation for each word by mapping the input words into embedding space, the \textit{hidden block} which includes one or more fully connected hidden layers with non-linear activation functions, and the \textit{softmax block} to generate a probability distribution over words. The main computational complexity of these NNLM are in the softmax and non-linear hidden blocks. 
CBOW and Skip-gram \cite{Mikolov_13-efficient, Mikolov_13-distributed} are two neural network architectures proposed to learn vector representations for words based on their neighboring contexts and with less computational complexity. A surprising aspect of the CBOW and Skip-gram models was that they were able to capture semantic similarities among word pairs with linear relationships in their vector space. GloVe \cite{Pennington_14-glove} was designed to capture explicitly what those methods do implicitly.

Given a sequence of $T$ training words $w_1, w_2, \dots, w_T$ from a corpus with the vocabulary $\bm{V}$ whose size is $|\bm{V}|$. Each model usually considers a context window of length $n$ and associates each word $w_t$ with an input embedding $\bm{v}_t$ and an output embedding $ \bm{v}^{\prime}_t$ which are vectors with dimension $d$. CBOW model aims at learning vector representations which are accurate at predicting a target word given $n$ words before and after the target word. If $\bm{\theta}$ denotes the parameters of the model, CBOW model maximizes: 
\begin{equation}
\label{eq:cbow}
J(\bm{\theta}) = \frac{1}{T} \sum_{t = 1}^{T} \log p(w_t | w_{t-n}, \dots , w_{t-1}, w_{t+1}, \dots, w_{t+n}) 
\end{equation}
where $J(\bm{\theta})$ denotes the objective function. On the other hand, the Skip-gram model learns the embedding representations which are good at predicting the surrounding terms of a given term. Mathematically, Skip-gram maximizes:
\begin{equation}
\label{eq:skipgram}
J(\bm{\theta}) = \frac{1}{T} \sum_{t = 1}^{T} \sum_{j = -n, \ne 0}^{n} \log p(w_{t+j} | w_t) 
\end{equation}
and the conditional probabilities are formulated using the word embeddings as:
\begin{equation}
p(w_{t+j} | w_t) = \frac{ \exp{( \bm{v}_{t+j}^{\prime T} \bm{v}_t})}{\sum_{i \in \bm{V}} \exp{( \bm{v}_{i}^{\prime T} \bm{v}_t)}}
\end{equation}

The global vectors for word representation in GloVe are learned to maintain the linear substructures in the vector space. The input is a matrix of $ |\bm{V}| \times |\bm{V}|$-dimension which is the word-word co-occurrence counts of word pairs rather than the entire corpus. To achieve this, a weighted  least squares objective function is designed as: 
\begin{equation}
\label{eq:glove}
J(\bm{\theta}) = \sum_{i, j \in \bm{V}} f(X_{ij}) (\bm{v}_{i}^{\prime T} \bm{v}_j + {b}_i + {b}_j^{\prime} - \log(X_{ij}))^2
\end{equation}
where $\bm{v}_{i}$ and $b_i$ are the vector representation and bias term for word $w_i$ and $\bm{v}_{j}^{\prime}$ and $b_j^{\prime}$ are the word representation and bias term for the context word $w_j$, $X_{ij}$ is the number of times $w_i$ occurs in the context of $w_j$ and $f$ is a weighting function that assigns relatively lower weights to rare and frequent terms.

%%%%%%%%%%%%%%%%%%%%%%%%%%%
\subsection{Score representation}
\label{ssec:score}
\textit{Score representation} as a low-dimensional feature set for sentiment classification was introduced in \cite{Farhadloo_13-multi}. Compared to classical Bag-Of-Words (BOW) features, score representation improved the performance of the sentiment identification task in aspect-level sentiment analysis and has been used to study customer satisfaction \cite{Farhadloo_16-modeling}. 

The sentiment identification as a classification task is a supervised machine learning task and requires manually labeled data. Score representation takes advantage of the given labeled training data in constructing its feature set. In a 3-class sentiment identification task, 3 scores are learned for each word $w_t$ in the vocabulary $|V|$ as:
\begin{align}
s^{+}_{t} & = \frac{f^{+}_{t}}{f^{+}_{t} + f^{0}_{t} + f^{-}_{t}}, \nonumber \\ 
s^{0}_{t} & = \frac{f^{0}_{t}}{f^{+}_{t} + f^{0}_{t} + f^{-}_{t}}, \\ 
s^{-}_{t} & = \frac{f^{-}_{t}}{f^{+}_{t} + f^{0}_{t} + f^{-}_{t}} \nonumber 
\end{align} 
where $f^{+}_{t}, f^{0}_{t}, f^{-}_{t}$ are the counts of term $w_t$ in positive, neutral and negative training documents respectively. In this approach, the scores are learned from the existing data (without using any external lexical resource) and reflect the positivity, neutrality and negativity of terms in the related content. Also in score representation, instead of working with $|V|$-dim vectors which usually is very large in a common text processing application, sentences can be represented with the 3-dimensional vectors which reflect the positiveness, neutralness and negativeness of sentences. It is worth mentioning that the approach expands easily to more than 3 classes as well. 
%%%%%%%%%%%%%%%%%%%%%%%%%%%%%%%%%%%%%%%%%%%%%%%%%%
\section{Methodology}
In this section we introduce the method of score embedding as a method capable of learning meaningful and interpretable vector representations for words. %To understand the method of score embedding first we give a brief description to the popular word embedding (WE) methods and also the score representation method for sentiment analysis.

%%%%%%%%%%%%%%%%%%%%%%%%%%%
\subsection{Score embedding architecture}
\label{ssec:scoreEmbed}
%%%%%%%%%%%%%%%%%%%%%%%%%%%%%%%%%%%%%%%%%%%%%%%%%%
\begin{figure*}
	\begin{center}
		\includegraphics[width = 2\columnwidth]{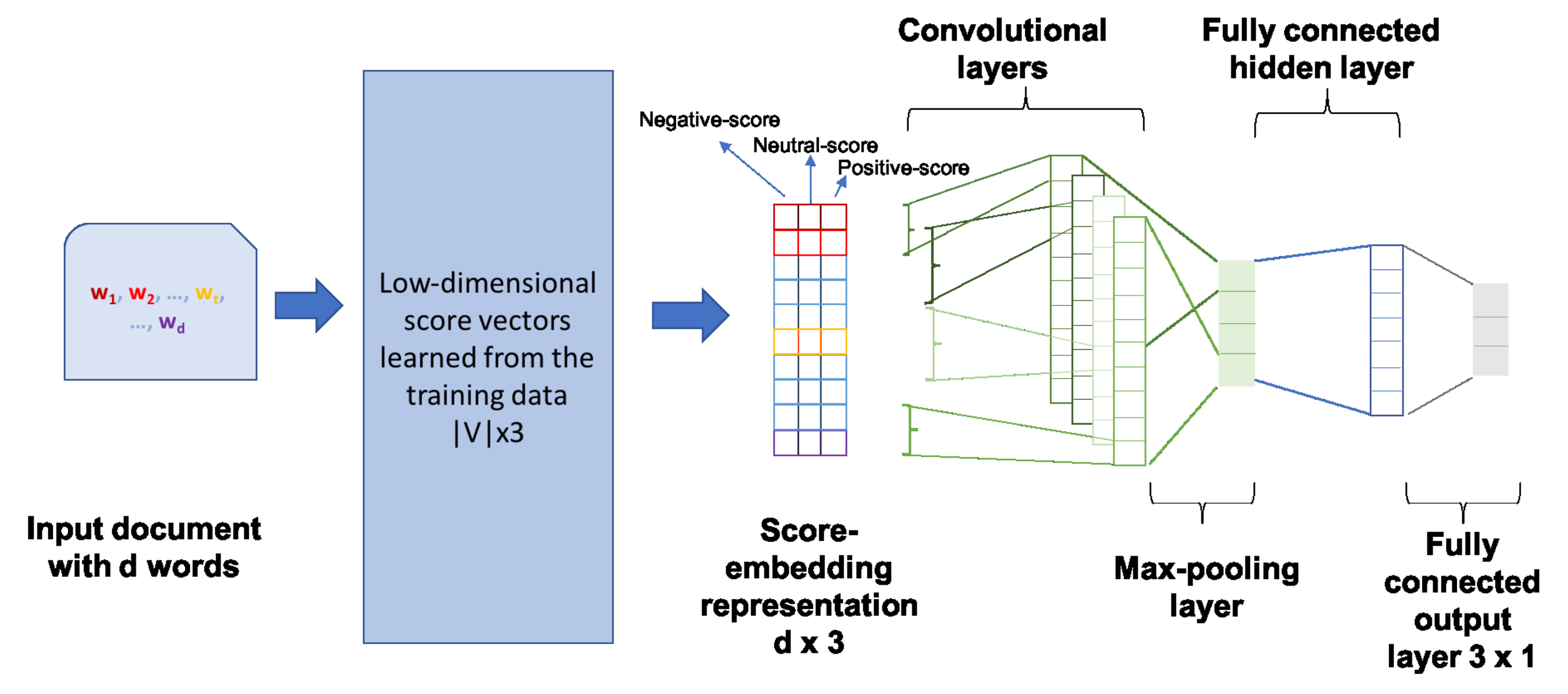} 
	\end{center}
	\caption{Score embedding architecture for a 3-class categorization task (it easily generalizes to C classes instead of 3). The input words are mapped into low-dimensional score embedding space and the neural network propagates the supervision from the output layer. The low-dimensional score vectors are initialized with the score representations learned from the labeled training data.}
	\label{fig:score}
\end{figure*}
%Figure \ref{fig:nnnlp} shows the architecture for a typical neural network language model. 
Many of the state-of-the-art Neural Network-based models for NLP tasks have a way to represent the input text documents. In these models, the low-dimensional embedding vectors can be pre-trained word vectors such as word2vec/GloVe, or be trained from scratch \cite{Collobert_11-natural, Socher_11-semiSupervised}. Learning pre-trained word vectors using the existing unsupervised techniques requires a huge amount of training documents which has a crucial influence on the performance of such systems \cite{Kim14-convolutional, JohnsonZhang14-effective, Gao_15-modeling}. The input representation of the model and the dimension of the embeddings are among the hyper-parameters of the model that have large impacts on the performance of the model on the tasks of interest.
%The performance of the neural network based methods for natural language processing tasks depend largely on the hyper-parameters of the model and the amount of available training data. One of the hyper-parameters is the input representation of the model. \cite{Kim14-convolutional} has experimented with different word vector representations such as word2vec or GloVe in a Convolutional Neural Network (CNN) architecture (Figure \ref{fig:nnnlp}). In \cite{JohnsonZhang14-effective, Gao_15-modeling} one-hot, BOW or bag of tri-letters vectors has been used to learn CNNs for text classification without the need to pre-trained word vectors. Many of the state-of-the-art papers that are using NN-based models for NLP tasks have a way to represent the input text documents. In Figure \ref{fig:nnnlp} the low-dimensional embedding vectors can be pre-trained word vectors such as word2vec/GloVe, or be trained from scratch. Learning pre-trained word vectors using the existing unsupervised techniques requires a huge amount of training documents which has a crucial impact on the performance of such systems. 
%%%%%%%%%%%%%%%%%%%%%%%%%%%%%%%%%%%
The architecture of the score embedding is depicted in Figure \ref{fig:score}. The input representation of the words in Figure \ref{fig:score} is based upon the scores learned from the provided labeled data instead of the one-hot, BOW, bag of tri-letters vectors, or pre-trained word vectors that have bee tried before \cite{Kim14-convolutional, JohnsonZhang14-effective, Gao_15-modeling}. The score representation, incorporated in score embedding architecture, efficiently and effectively leverages the class information provided through labeled data.

Let $\bm{x}_t \in \mathbb{R}^{C}$ be the C-dimensional score vector corresponding to the $t^{th}$ word in the sentence. A sentence of length $n$ is represented by concatenating the vectors of the its constituent words selected from the $|V| \times C$ matrix of score vectors (shorter sentences should be padded if necessary). The low-dimensional score vectors in Figure \ref{fig:score} are learned from the labeled training data as described in section \ref{ssec:score}. A convolution operation generates a new feature $c_t$ by applying a filter $\bm{w} \in \mathbb{R}^{hC}$ to a window of $h$ words $\bm{x}_{t:t+h-1}$:
\begin{equation}
c_t = a(\bm{w} \cdot \bm{x}_{t:t+h-1} + b)
\end{equation}
where $b$ is a scalar bias term and $a(.)$ is the activation function which could be non-linear. The convolutional layer applies the operation to each possible sequence of words in the sentence and produces a feature map $\bm{c} \in \mathbb{R}^{n-h+1}$ by concatenating them together. To deal with the variable sentence lengths and also capture the most important feature, a max-over-time pooling operation is applied to each feature map. The model in Figure \ref{fig:score} uses multiple filters with varying window sizes to produce multiple features. These features are then passed to a fully connected softmax layer whose output is the probability of each category.
%%%%%%%%%%%%%%%%%%%%%%%%%%%%%%%%%%%
\subsection{Embedding dimension and initialization}
\label{ssec:dimension}
In existing embedding methods, the dimension which the words are mapped in the low dimensional space is one of the hyper-parameters of the model and is usually determined by cross-validating values in the range of $50-600$. However, not many of the unsupervised or supervised methods have paid attention to the meanings and the interpretations of their representations in the vector space as achieving some continuous representations whose dimension is lower than the BOW is their objective. 

Score embedding addresses this shortcoming by learning meaningful word vectors based on the score representation described in section \ref{ssec:score} and the supervision provided with the labeled training data. First, instead of initializing the word vectors with those obtained from an unsupervised neural language model, we consider the scores learned for each word based on the score representations as pre-trained embeddings. Second, in the classification task, the scores of each term are learned using the provided training data through back-propagation. Each score in our model indicates the association of each term with the existing classes. If there are $C$ categories, we embed the word $w_t$ into a $C$ dimensional vector space $\bm{x}_t = [x_{t1}, \dots, x_{tC}]$ where $x_{tc} = s_t^c$ is the score representation of the $c^{th}$ category of the $t^{th}$ word. For instance, in a 3-class sentiment classification task, the 3 scores of each term are respectively the positiveness, neutralness and negativeness of each term in the particular domain of interest. In the score embedding the learned scores are considered as pre-trained word vectors and are fine tuned using the neural network architecture.

%%%%%%%%%%%%%%%%%%%%%%%%%%%%%%%%%%%
\subsection{Learning the parameters of the model}
\label{ssec:learn}
If $\bm{\theta}$ denotes all the trainable parameters of the network in Figure \ref{fig:score}, score embedding maximizes the log-likelihood of the training data with respect to $\bm{\theta}$:
%%%%%%%%%%%%%%%%%%
\begin{equation}
\label{eq:score}
J(\bm{\theta}) = \sum_{(\bm{x},y) \in \bm{D}} \log p(y | \bm{x}, \bm{\theta})
\end{equation}
%%%%%%%%%%%%%%%%%
where $\bm{x}$ corresponds to the training word associated features, $y$ represents the corresponding output label, and $\bm{D}$ is the set of all training examples. The conditional probability of $p(y | (\bm{x}, \bm{\theta}))$ is calculated using the outputs of the network. If the corresponding network output for the $c^{th}$ class is denoted as $f_{\bm{\theta}}^{c}(\bm{x})$ then: 
\begin{equation}
p(y = c | \bm{x}, \bm{\theta}) = \frac{\exp{f_{\bm{\theta}}^{c}(\bm{x})}}{\sum_{j} \exp{f_{\bm{\theta}}^{j}(\bm{x})}}
\end{equation}
and the log-likelihood for one training example $(\bm{x},y)$ is:
\begin{equation}
\log p(y = c | \bm{x}, \bm{\theta}) = f_{\bm{\theta}}^{c}(\bm{x}) - \log \sum_{j} \exp{f_{\bm{\theta}}^{j}(\bm{x})}
\end{equation}
The objective function, $J(\bm{\theta})$, in equation \ref{eq:score} can be maximized by taking the derivative of it through back-propagation with respect to the whole set of parameters, and use Ada-Grad \cite{Duchi_11-adaptive} to update the parameters.

%%%%%%%%%%%%%%%%%%%%%%%%%%%%%%%%%%%%%%%%%%%%%%%%%%
\section{Experiments}
%%%%%%%%%%%%%%%%%%%%%%%%%%%
\subsection{Data}
We have used two data sets for evaluating our method. The first data set is the \textbf{ACA data} which we have collected to study the public opinions about the health-care on Twitter. The second data is the Stanford Sentiment Treebank (SST) which is used as a benchmark to be able to compare our results with other studies in the literature.

\paragraph{ACA data} Given  a set of keywords related to affordable care act (``ACA", ``Obamacare", ``affordable care act", ``health care", ``healthcare", ``ryancare", ``single payer", ``singlepayer", ``trumpcare", ``insured", ``uninsured", ``savetheACA", ``saveaca", ``congress", ``vote", ``WorldsGreatestHealthCarePlan", ``HR1275", ``H.R.  1275"), tweets were collected using Twitter streaming API from March, 2016 to July, 2016. Not all of the collected tweets were related to the discussions of affordable care act on Twitter  as some keywords such as ``vote" or ``congress" were ambiguous. After removing the unrelated tweets by enforcing the existence of one of the mentioned keywords and ``health care", there were 45\,847\,642 tweets in our corpus. To prepare a training/testing set, $3000$ tweets were randomly selected and manually labeled by 2 annotators considering 4 different classes of: ``against TrumpCare", ``neutral", ``pro TrumpCare", or ``do not know/not related" (the Cohen's kappa coefficient to measure the agreement between them was $\%0.87$). The statistics of the labeled training/testing data is summarized in Table \ref{tab:training}. There were 542 ``do not know/not related" tweets which were discarded from the sets.

\paragraph{Stanford Sentiment Treebank (SST)} SST \cite{Socher_13-recursive} has been used as a benchmark dataset for sentiment classifications. It contains movie review snippets with fine-grained labels of \textit{very negative, negative, neutral, positive and very positive} with train/dev/test splits. It has provided train/dev/test sets with 8544/1101/2210 samples, respectively.
%%%%%%%%%%%%%%%%%%%%%%%%%%%%%%%%%%%%%%%%
\begin{table*}
	\small
	\begin{center}
		\caption{\small{Training and Testing statistics of the ACA data}}
		\label{tab:training}
		\begin{tabular}{lccc}
			\hline
			& against TrumpCare & neutral & pro TrumpCare \\
			\hline
			Training & 1216 & 676 & 320 \\
			Testing & 135 & 76 &  35\\
			Total & 1351 & 752 & 355  \\
			\hline
		\end{tabular}
	\end{center}
\end{table*}
%%%%%%%%%%%%%%%%%%%%%%%%%%%
\subsection{Results}
In this section we report the performance of different methods that was measured using 5-fold cross-validation. In the score embedding experiments, the network consisted of a convolutional layer with 128 filters of sizes 3, 4 and 5 words. Dropout was also employed as a regularization technique to avoid over-fitting.
%%%%%%%%%%%%%%%%%%%%%%%%%%%
\paragraph{Lexicon-based methods} Many of the existing sentiment analyses studies are based on a sentiment lexicons \cite{Hu_04-mining,Baccianella_10-sentiwordnet, Mohammad_13-nrc}. Opinion Miner (OM) lexicon \cite{Hu_04-mining} and SentiWordNet \cite{Baccianella_10-sentiwordnet} are among the popular sentiment lexicons. OM lexicon consists of 2006 positive and 4783 negative terms which also includes common misspellings, morphological variants, slangs and social media mark-ups. SentiWordNet assigns an objectivity score and a positive and negative score to each WordNet \cite{Miller95-wordnet} synset in that these three scores sum up to 1. To evaluate the performance of the lexicon based methods, we applied these lexicons to our test set and calculated the accuracy of them. A given sentence is positive (negative), if the number of positive terms is greater than the number of negative (positive) terms and is neutral if they are equal (Table \ref{tab:acaResult}). 
%%%%%%%%%%%%%%%%%%%%%%%%%%%
\paragraph{Classic supervised methods} The classic supervised methods mainly use bag of words as their features. Here we have trained Naive Bayes, Random Forrest and Support Vector Machines with linear, polynomial and Radial Basis Functions (RBF) kernels. We have considered linear and nonlinear classifiers to have a fair comparison with the neural network based methods as some studies in the literature when compare the classic techniques with neural network models, do not do a extensive parameter tuning for the classic methods compared to their proposed methods \cite{Socher_13-recursive,JohnsonZhang14-effective}. Tables \ref{tab:acaResult} and \ref{tab:sstResult} summarizes the results.
%%%%%%%%%%%%%%%%%%%%%%%%%%%
\paragraph{Word-embedding methods} %Initializing word vectors with those obtained from an unsupervised neural language model is a popular method to improve performance in the absence of a large supervised training set \cite{Collobert_11-natural, Socher_11-semiSupervised}. 
We used the sentiment specific word embedding (SSWE) \cite{Tang_14-sswe} which were trained using distance supervision particularly to capture sentiment information as a comparable method. We also used the publicly available embedding vectors \cite{Pennington_14-glove} trained using either Twitter or Common Crawl data in our comparisons.
%We have also used the sentiment specific word embedding (SSWE) \cite{Tang_14-sswe} which were trained using distance supervision particularly to capture sentiment information as another comparable method. 
The word vectors pre-trained on Twitter were available in dimensions 25, 50, 100, 200 and were trained on 27 billion tokens from 2 billion tweets with the vocabulary size of 1.2 million. The embeddings pre-trained on Common Crawl data were only available in dimension 300 and were trained on 840 billion tokens with vocabulary size of 2.2 million. The pre-trained word vectors were used as features and SVM classifiers with polynomial and Radial Basis Functions (RBF) kernels were designed for performing 3-class sentiment analysis on the ACA data. Since it is highly recommended to normalize the features for SVM classifiers, we have conducted experiments with and without normalization. %(Figure \ref{fig:pretrained}). 
%%%%%%%%%%%%%%%%%%%%%%%%%%%
\begin{figure*}
	\begin{tabular}{lccc}
		\raisebox{4ex}{\rotatebox{90}{Accuracy}} & \includegraphics[width=0.3\textwidth]{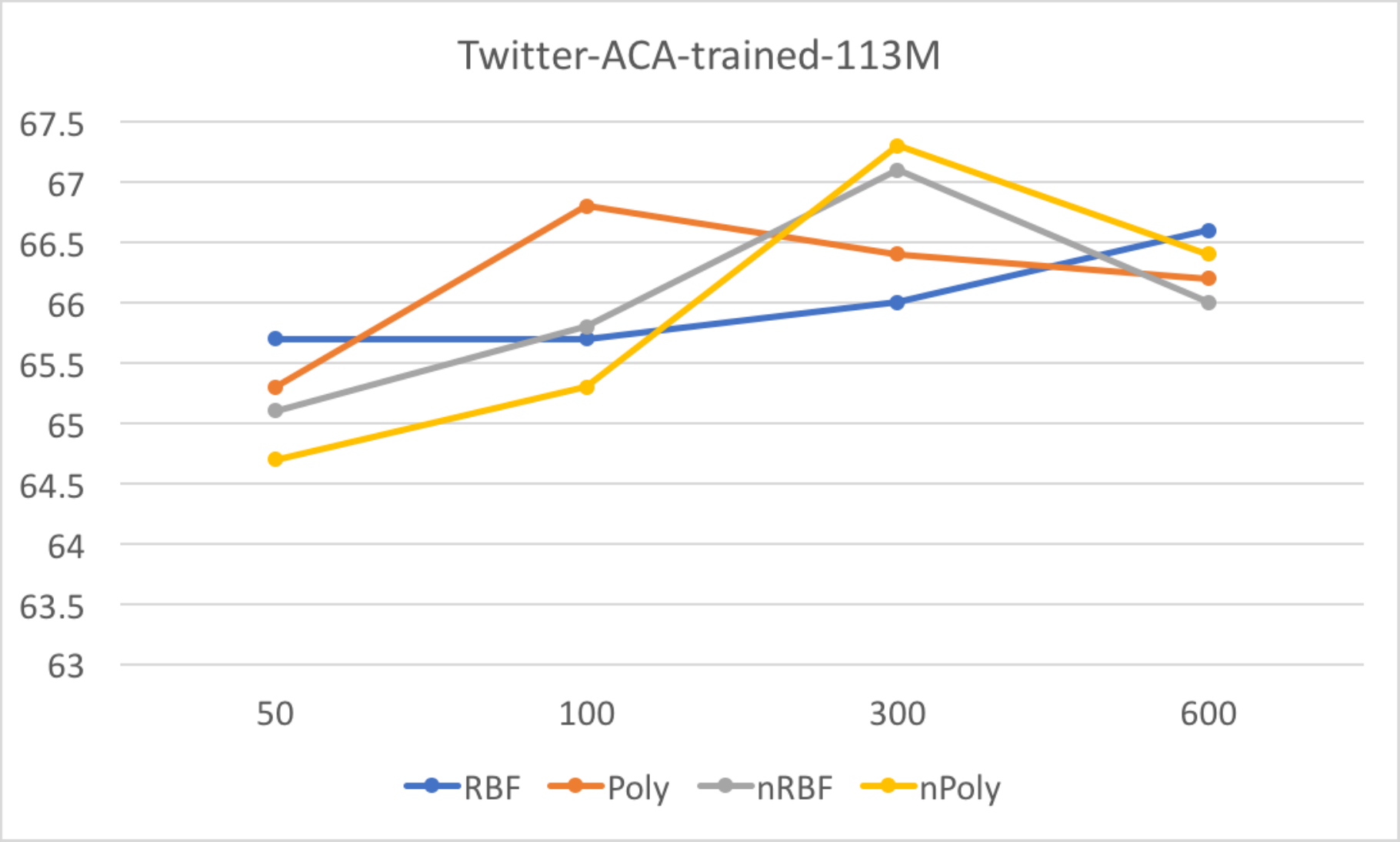} &
		\includegraphics[width=0.3\textwidth]{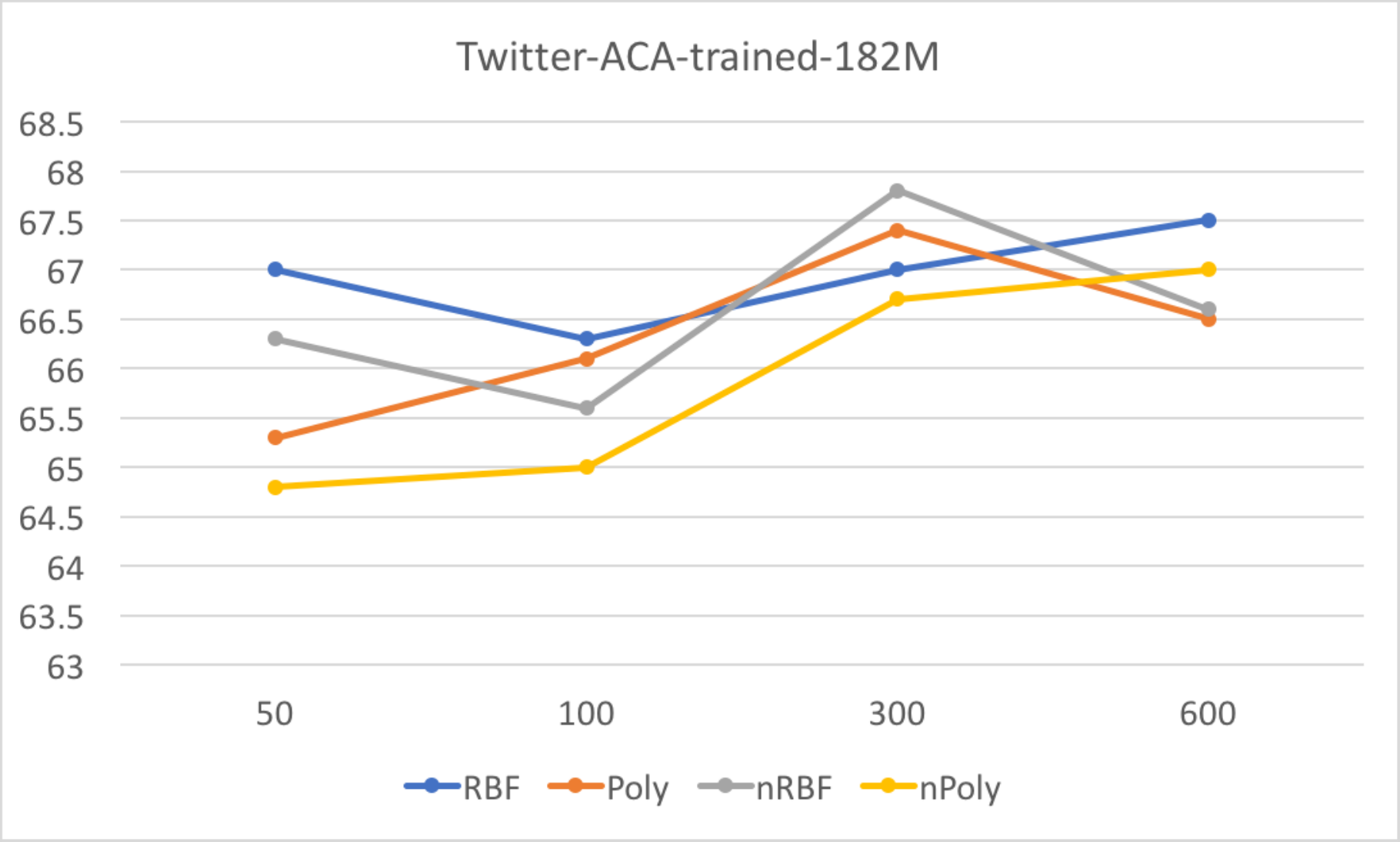} &
		\includegraphics[width=0.3\textwidth]{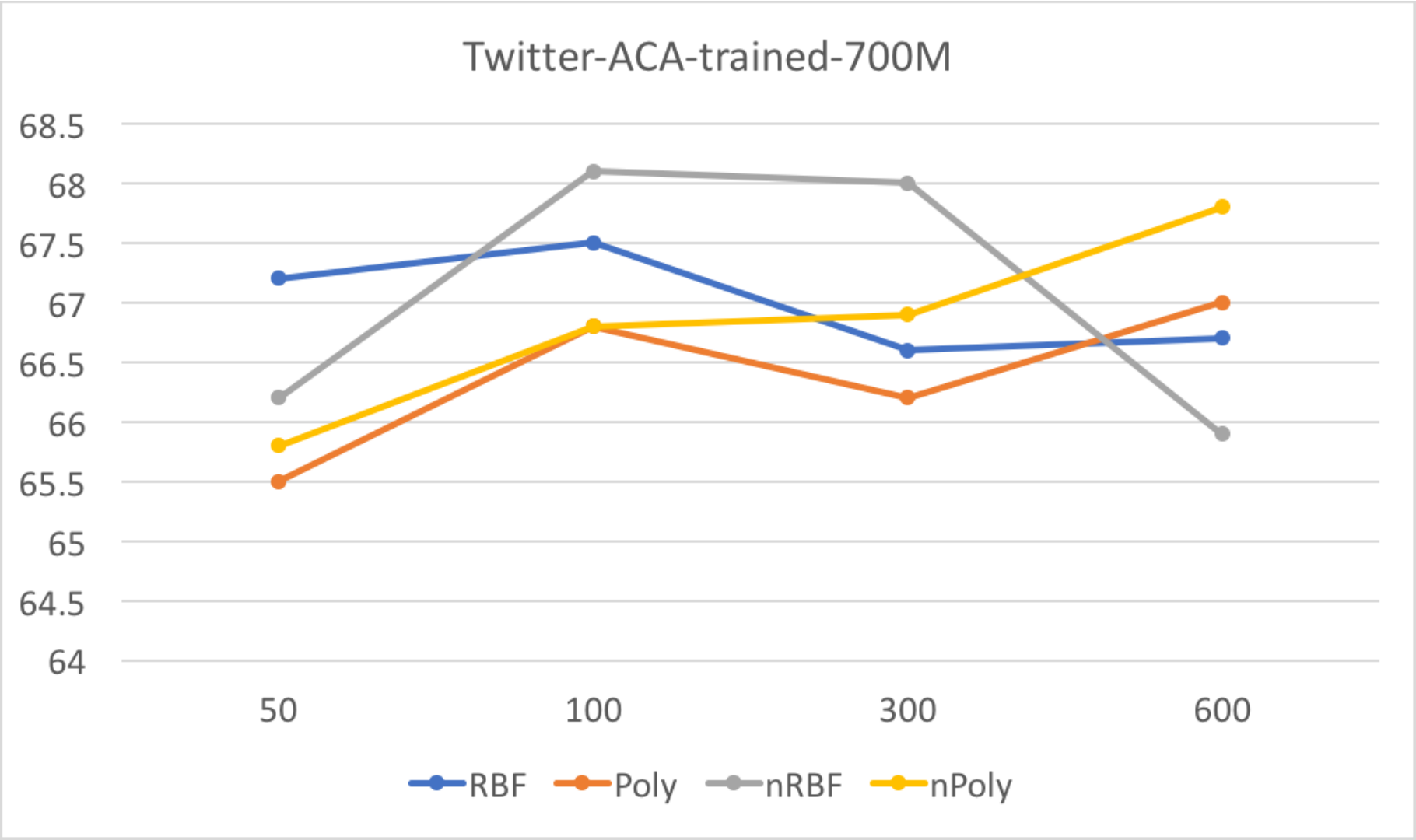} \\
		& Embedding dimension & Embedding dimension & Embedding dimension \\
		& (a) & (b) & (c)
	\end{tabular}
	\caption{3-class sentiment analyses using the word embeddings trained on ACA data as features and radial basis function (RBF) and polynomial (poly) kernels. For nRBF and nPoly the features were normalized first. Embeddings were trained on (a) 25\% of the data with 113 million tokens, (b) 50\% of the data with 182 million tokens, and (c) all of the ACA data with 700 million tokens.}
	\label{fig:trained}
\end{figure*}	
%%%%%%%%%%%%%%%%%%%%%%%%%%%
%%%%%%%%%%%%%%%%%%%%%%%%%%%%%%%%%%%%
To customize the word vectors to our ACA data, we also trained the word vectors using the ACA data and investigated the impact of the available training data on the performance of them. We prepared 3 versions of our ACA data to train the word vectors: 
\vspace{-2ex}
\begin{description}
	\item[Twitter-ACA-113M:] $25\%$ of our ACA data (randomly selected) which has 113 million tokens.
	\vspace{-2ex}
	\item[Twitter-ACA-182M:] $50\%$ of our ACA data (randomly selected) which has 182 million tokens.
	\vspace{-2ex}
	\item[Twitter-ACA-700M:] all of the ACA data with 700 million tokens.
\end{description}
\vspace{-2ex}
Again SVM classifiers with polynomial and Radial Basis Functions (RBF) kernels were designed for performing 3-class sentiment analysis on the ACA data using the trained word vectors as features. Since it is highly recommended to normalize the features for SVM classifiers, we have conducted experiments with normalization and without it. 

The unsupervised word vectors trained on a large data set with 840 billion tokens outperformed the supervised SSWE method. One reason for this could be the distance supervision of the SSWE provided by emoticons is noisy. Also we observed that for the pre-trained word vectors the RBF kernels outperform the polynomial kernels and there is no need to normalize the pre-trained word vectors. Consistent with other studies we also found that the amount of available unlabeled training data had an impact on the performance as the word vectors from the Common Crawl data with 840 billion tokens gave better results than the Twitter data with 2 billion tokens. %(Figure \ref{fig:pretrained}). 

Training the word vectors from scratch using the ACA data (Figure \ref{fig:trained}) also validates that the word vectors trained on a larger data set provide better features for the subsequent categorization task. Our findings also show that the embedding dimension impacts on the performance but it is not always the case that increasing the embedding dimension will improve the categorization results. Indeed, we achieved the best performance (Figure \ref{fig:trained} (c)) with the embedding dimension of 100 and normalized features for the RBF kernels.
%%%%%%%%%%%%%%%%%%%%%%%%%%%
\vspace{-2ex}
\paragraph{Score representation and Score embedding} Using the available ACA training data, scores were learned for each term in our vocabulary. The learned scores then were used as features as described in section \ref{ssec:score} for SVM classification. The learned scores were also used as pre-trained word vectors to initialize the low-dimensional word-embeddings in Figure \ref{fig:score} and then the neural architecture learns score embedding through fine tuning using the sentiment supervision. 

%%%%%%%%%%%%%%%%%%%%%%%%%%%
\begin{table*}
	\centering
	\small{
		\begin{tabular}{ccccccccccccc}
			\hline
			& \multicolumn{2}{c|}{Lexicon-based} & \multicolumn{4}{c|}{Classic-supervised} & \multicolumn{3}{c|}{Unsupervised WE} & \multicolumn{3}{c}{Supervised} \\ \hline
			& OM lex. & Senti. & NB & RF & lSVM & rbfSVM & Paragh-vec & WE (pre-trained) & WE & SSWE & Score & SE \\ \hline
			Accuracy & 37 & 31 & 59 & 66.3 & 66.6 & 66.8 & 65.0 & 66.5 & 68.1 & 60.2 & 68.3 & 69.4 \\ \hline
	\end{tabular}}
	\caption{3-class sentiment analysis on Affordable Care Act data with score embedding and other methods: \textbf{OM lex.}  \textbf{Senti.}, \textbf{NB:} Naive Bayes with BOW, \textbf{RF:} random forest classifier with BOW, \textbf{lSVM:} linear SVM, \textbf{rbfSVM:} SVM with radial basis functions, \textbf{Paragh-vec:} \cite{Le_14-distributed}, \textbf{WE:}  pre-trained and trained word vectors, \textbf{SSWE:}  \cite{Tang_14-sswe}, \textbf{score:} score representation as features with SVM classifiers, and \textbf{SE:} score embedding.}
	\label{tab:acaResult}
\end{table*}
%%%%%%%%%%%%%%%%%%%%%%%%%%%
\begin{table*}
	\centering
	\small{
		\begin{tabular}{cccccccccccc}
			\hline
			& \multicolumn{3}{c|}{Classic-supervised} & \multicolumn{5}{c|}{Unsupervised WE features} & \multicolumn{3}{c}{Supervised} \\ \hline
			& NB & SVM & BiNB & RAE & MV-RNN & RNTN & CNN & Paragh-vec & SSWE & Score & SE \\ \hline
			Accuracy & 41.0 & 40.7 & 41.9 & 43.2 & 44.4 & 45.7 & 45.0 & 48.7 & 33.6 & 45.0 & 46.0 \\ \hline
	\end{tabular}}
	\caption{5-class sentiment analysis on Stanford Sentiment Treebank with Score embedding and other methods: \textbf{NB:} Naive Bayes, \textbf{SVM:}  \cite{Socher_13-recursive}, \textbf{BiNB:} Naive Bayes with bag of bi-grams as features \cite{Socher_13-recursive}, \textbf{RAE:} \cite{Socher_11-semiSupervised}, \textbf{MV-RNN:} \cite{Socher_12-semantic}, \textbf{RNTN:} \cite{Socher_13-recursive}, \textbf{CNN:} \cite{Kim14-convolutional}, \textbf{Paragh-vec:} \cite{Le_14-distributed}, \textbf{SSWE:}, \textbf{score:} score representation with SVM classifiers and \textbf{SE:} score embedding.}
	\label{tab:sstResult}
\end{table*}
%%%%%%%%%%%%%%%%%%%%%%%%%%%

In Tables \ref{tab:acaResult} and \ref{tab:sstResult} the score representation method which is a supervised feature engineering method for sentiment analysis generates comparable results as the unsupervised word embedding techniques which generally require a large amount of data. Score embedding which incorporates both score representation and neural network architecture outperforms it by effectively using the labeled data. It is worth mentioning that although score embedding does not outperform the paragh-vec method in Table \ref{tab:sstResult}, it is competitive to many of the state-of-the-art methods. On the other hand, score embedding is not computationally expensive and relatively simple unlike paragh-vec which suffers from large processing time even at the testing \cite{Le_14-distributed}. Overall, our results show that score embedding, a relatively simple approach, which combines score representation with neural network architecture, can perform at the state-of-the-art level, even performing competitively with methods rely on huge amount of learning data.

\begin{figure*}
	\begin{center}
		\includegraphics[width = 2\columnwidth]{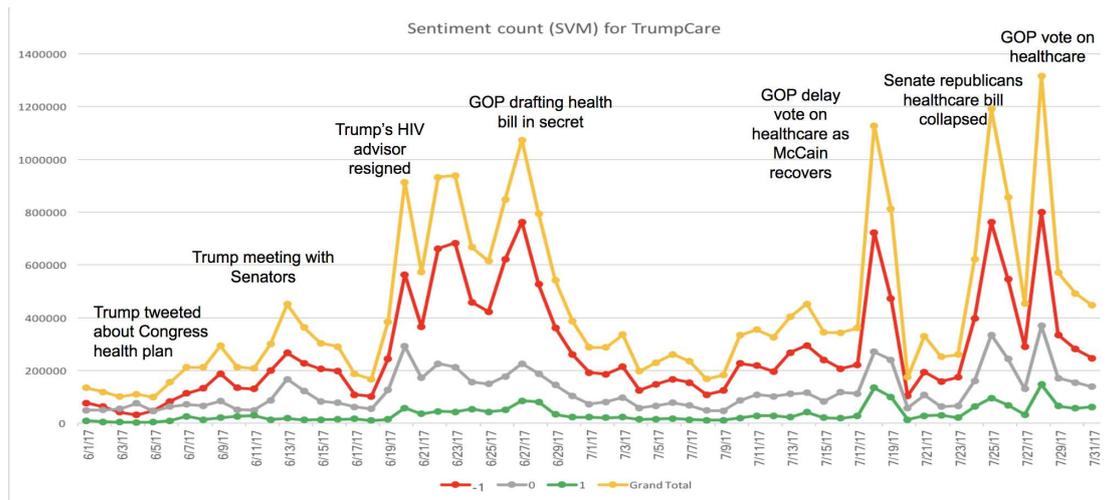} 
	\end{center}
	\caption{Sentiment analysis of the affordable care act data over time. The negative sentiment towards ``TrumpCare'' was more than neutral and positive sentiments. At each day, total number of tweets and number of negative (-1), neutral (0), and positive (+1) sentiments have been shown.}
	\label{fig:aca}
\end{figure*}

The score embedding classifier was applied to a subset of our ACA data from June to July 2017 with 26\,604\,224 tweets. After identifying the sentiment orientation of each tweet towards ``TrumpCare", then we aggregated the volume of each category over time (Figure \ref{fig:aca}). Interestingly our results show that, the negative sentiment towards ``TrumpCare" consistently was greater than the neutral and positive sentiment over time. It is worth noting that such analyses can be used to find a fine-grained timeline of the affordable care act as we were able to find a related event corresponding to the peaks (valleys) of Figure \ref{fig:aca}.
%%%%%%%%%%%%%%%%%%%%%%%%%%%%%%%%%%%

%%%%%%%%%%%%%%%%%%%%%%%%%%%
\section{Conclusion}
The existing unsupervised word embeddings do not effectively capture the sentiment information and the dimension to which the words are mapped is a hyper-parameter of the model which impacts the performance. In this paper we proposed score embedding as a supervised approach to learn vector representations which incorporates the supervision provided with labeled training data in two ways. First, the supervision is incorporated in initializing the word vectors using the scores learned for each word from the labeled data. Second, the neural architecture fine-tunes the embeddings by maximizing the log-likelihood of the output categories over the word vectors. Our results show that score embedding can perform at the state-of-the-art level, even performing competitively with methods rely on huge amount of learning data.

\bibliography{MF-conll2018}
\bibliographystyle{plain}

\appendix

\end{document}